\documentclass[journal]{IEEEtran}

\usepackage{cite}

\ifCLASSINFOpdf
  \usepackage[pdftex]{graphicx}
\else
\fi

\usepackage{url}
\usepackage{multirow}
\usepackage{amsmath}
\usepackage{amssymb}		
\usepackage{siunitx}
\usepackage{booktabs} 	

\newcommand{\dataset}{{\cal D}}

\begin{document}

\title{Integration of Neural Network-Based Symbolic Regression in Deep Learning for Scientific Discovery}

\author{Samuel~Kim$^{1,a}$, Peter~Lu$^2$, Srijon~Mukherjee$^2$, Michael~Gilbert$^1$, Li~Jing$^2$, Vladimir~\v{C}eperi\'{c}$^3$, Marin~Solja\v{c}i\'{c}$^2$
\thanks{$^1$Department of Electrical Engineering and Computer Science, Massachusetts Institute of Technology, Cambridge, MA, USA}
\thanks{$^2$Department of Physics, Massachusetts Institute of Technology, Cambridge, MA, USA}
\thanks{$^3$Faculty of Electrical Engineering and Computing, University of Zagreb, Zagreb, Croatia}
\thanks{$^a$E-mail: samkim@mit.edu}}

\maketitle

\begin{abstract}
Symbolic regression is a powerful technique that can discover analytical equations that describe data, which can lead to explainable models and generalizability outside of the training data set. In contrast, neural networks have achieved amazing levels of accuracy on image recognition and natural language processing tasks, but are often seen as black-box models that are difficult to interpret and typically extrapolate poorly. Here we use a neural network-based architecture for symbolic regression called the Equation Learner (EQL) network and integrate it with other deep learning architectures such that the whole system can be trained end-to-end through backpropagation. To demonstrate the power of such systems, we study their performance on several substantially different tasks. 
First, we show that the neural network can perform symbolic regression and learn the form of several functions. Next, we present an MNIST arithmetic task where a separate part of the neural network extracts the digits. Finally, we demonstrate prediction of dynamical systems where an unknown parameter is extracted through an encoder. We find that the EQL-based architecture can extrapolate quite well outside of the training data set compared to a standard neural network-based architecture, paving the way for deep learning to be applied in scientific exploration and discovery.
\end{abstract}

\begin{IEEEkeywords}
Symbolic Regression, Neural Network, Kinematics, Simple Harmonic Oscillator, ODE, Discovery
\end{IEEEkeywords}

\IEEEpeerreviewmaketitle

\section{Introduction}

Many complex phenomena in science and engineering can be reduced to general models that can be described in terms of relatively simple mathematical equations. For example, classical electrodynamics can be described by Maxwell's equations and non-relativistic quantum mechanics can be described by the Schr\"{o}dinger equation. These models elucidate the underlying dynamics of a particular system and can provide general predictions over a very wide range of conditions. On the other hand, modern machine learning techniques have become increasingly powerful for many tasks including image recognition and natural language processing, but the neural network-based architectures in these state-of-the-art techniques are black-box models that often make them difficult for use in scientific exploration. In order for machine learning to be widely applied to science, there is a need for interpretable and generalizable models that can extract meaningful information from complex datasets and extrapolate outside of the training dataset.

Symbolic regression is a type of regression analysis that searches the space of mathematical expressions to find the best model that fits the data. 
It is much more general than linear regression in that it can potentially fit a much wider range of datasets and does not rely on a set of predefined features. Assuming that the resulting mathematical expression correctly describes the underlying model for the data, it is easier to interpret and can extrapolate better than black-box models such as neural networks. 
Symbolic regression is typically carried out using techniques such as genetic programming, in which a tree data structure representing a mathematical expression is optimized using evolutionary algorithms to best fit the data \cite{Koza1994}. Typically in the search for the underlying structure, model accuracy is balanced with model complexity to ensure that the result is interpretable and does not overfit the data.
This approach has been used to extract the underlying laws of physical systems from experimental data \cite{Schmidt2009}. However, due to the combinatorial nature of the problem, genetic programming does not scale well to large systems and can be prone to overfitting.

Alternative approaches to finding the underlying laws of data have been explored. For example, sparsity has been combined with regression techniques and numerically-evaluated derivatives to find partial differential equations (PDEs) that describe dynamical systems \cite{Brunton2016,Rudy2017,Schaeffer2017}. 

There has also been significant work on designing neural network architectures that are either more interpretable or more applicable to scientific exploration. Neural networks with unique activation functions that correspond to functions common in science and engineering have been used for finding mathematical expressions that describe datasets \cite{Martius2016,Sahoo2018}. A deep learning architecture called the PDE-Net has been proposed to predict the dynamics of spatiotemporal systems and produce interpretable differential operators through constrained convolutional filters \cite{Long2018,Long2018-2}. \cite{Trask2018} propose a neural network module called the Neural Arithmetic Logic Unit (NALU) that introduces inductive biases in deep learning towards arithmetic operations so that the architecture can extrapolate well on certain tasks. 
Neural network-based architectures have also been used to extract relevant and interpretable parameters from dynamical systems and use these parameters to predict the propagation of a similar system \cite{Zheng2018, Lu2019}. Additionally, \cite{Chari2019} use symbolic regression as a separate module to discover kinematic equations using parameters extracted from videos of balls under various types of motion.

Here we present a neural network architecture for symbolic regression that is integrated with other deep learning architectures so that it can take advantage of powerful deep learning techniques while still producing interpretable and generalizable results. Because this symbolic regression method can be trained through backpropagation, the entire system can be trained end-to-end without requiring multiple steps.

Source code is made publicly available \footnote{\url{https://github.com/samuelkim314/DeepSymReg}}.

\section{EQL Architecture}

\begin{figure}[t]
	\centering
    \includegraphics[width=0.9\columnwidth]{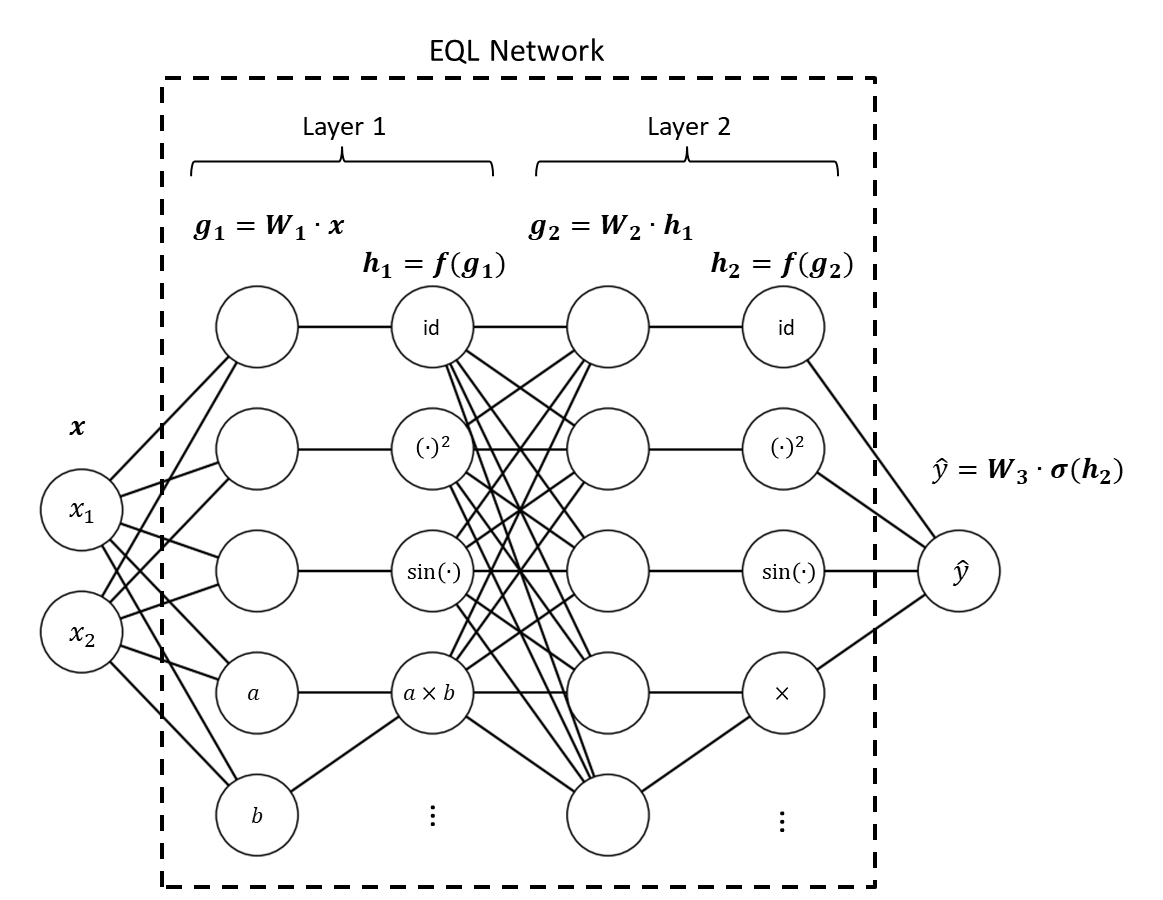}
	\caption{Example of the Equation Learner (EQL) network for symbolic regression using a neural network. Here we show only 4 activation functions (identity or ``id", square, sine, and multiplication) and 2 hidden layers for visual simplicity, but the network can include more functions or more hidden layers to fit a broader class of functions.}
	\label{fig:eql}
\end{figure}

The symbolic regression neural network we use is similar to the Equation Learner (EQL) network proposed in \cite{Martius2016,Sahoo2018}. As shown in Figure \ref{fig:eql}, the EQL network is based on a fully-connected neural network where the $i^{\mathrm{th}}$ layer of the neural network is described by 
\begin{align*}
\mathbf{g}_i & = \mathbf{W}_i\mathbf{h}_{i-1} \\
\mathbf{h}_i & = f\left(\mathbf{g}_i \right)
\end{align*}
\noindent where $\mathbf{W}_i$ is the weight matrix of the $i^{\mathrm{th}}$ layer and $\mathbf{h}_0=\mathbf{x}$ is the input data. The final layer does not have an activation function, so for a network with $L$ hidden layers, the output of the network is described by
\[ y=\mathbf{h}_{L+1}=\mathbf{W}_{L+1}\mathbf{h}_{L} \]

The activation function $f(\mathbf{g})$, rather than being the usual choices in neural networks such as ReLU or tanh, may consist of a separate function for each component of $\mathbf{g}$ (such as sine or the square function) and may include functions that take two or more arguments while producing one output (such as the multiplication function):
\begin{equation}
 f(\mathbf{g})= \begin{bmatrix} f_1(g_{1}) \\ f_2(g_{2}) \\ \vdots \\ f_{n_h}(g_{n_g-1},g_{n_g}) \end{bmatrix}
\label{eq:activation}
\end{equation}
\noindent Note that an additive bias term can be absorbed into $f(\mathbf{g})$ for convenience. These activation functions in (\ref{eq:activation}) are analogous to the primitive functions in symbolic regression. Allowing functions to take more than one argument allows for multiplicative operations inside the network. 

While the schematic in Figure \ref{fig:eql} only shows 4 activation functions in each hidden layer for visual simplicity, $f(\mathbf{g})$ in \ref{eq:activation} can include other functions including $\exp{(g)}$ and $\text{sigmoid}(g)=\frac{1}{1+e^{-g}}$. Additionally, we allow for activation functions to be duplicated within each layer. This reduces the system's sensitivity to random initializations and creates a smoother optimization landscape so that the network does not get stuck in local minima as easily. This also allows the EQL network to fit a broad range of functions. More details can be found in Appendix \ref{app:benchmark}.

By stacking multiple layers (i.e. $L\geq 2$), the EQL architecture can fit complex combinations and compositions of a variety of primitive functions. $L$ is analogous to the maximum tree depth in genetic programming approaches and sets the upper limit on the complexity of the resulting expression. 
While this model is not as general as conventional symbolic regression, it is powerful enough to represent most of the functions that are typically seen in science and engineering. More importantly, because the EQL network can be trained by backpropagation, it can be integrated with other neural network-based models for end-to-end training.

\subsection{Sparsity}

A key ingredient of making the results of symbolic regression interpretable is enforcing sparsity such that the system finds the simplest possible equation that fits the data. The goal of sparsity is to set as many weight parameters to 0 as possible such that those parameters are inactive and can be removed from the final expression. Enforcing sparsity in neural networks is an active field of research as modern deep learning architectures using millions of parameters start to become computationally prohibitive \cite{Louizos2017,Molchanov2017,Zhu2017}. \cite{Gale2019} evaluates several recent developments in neural network sparsity techniques. 

A straightforward and popular way of enforcing sparsity is adding a regularization term to the loss function that is a function of the neural network weight matrices:
\begin{equation}
L_q=\sum_{i=0}^{L+1}\left\lVert\mathbf{W}_i\right\rVert^q
\label{eq:reg}
\end{equation}
\noindent where $\left\lVert\mathbf{W}_i\right\rVert^q$ is the element-wise norm of the matrix:
\[ \left\lVert\mathbf{W}_i\right\rVert^q = \sum_{j,k} \left| w_{j,k} \right|^q  \]

Setting $q=0$ in (\ref{eq:reg}) results in $L_0$ regularization, which penalizes weights for being non-zero regardless of the magnitude of the weights and thus drives the solution towards sparsity. However, $L_0$ regularization is equivalent to a combinatorics problem that is NP-hard, and is not compatible with gradient descent methods commonly used for optimizing neural networks \cite{Natarajan1995}. Recent works have explored training sparse neural networks with a relaxed version of $L_0$ regularization through stochastic gate variables, allowing this regularization to be compatible with backpropagation \cite{Louizos2017,Srinivas2017}. 

A much more popular and well-known sparsity technique is $L_1$ regularization, which is used in the original EQL network \cite{Martius2016}. Although it does not push solutions towards sparsity as strongly as $L_0$ regularization, $L_1$ regularization is a convex optimization problem that can be solved using a wide range of optimization techniques including gradient descent to drive the weights towards 0. However, while $L_1$ is known to push the solution towards sparsity, it has been suggested that $L_{0.5}$ enforces sparsity more strongly without penalizing the magnitude of the weights as much as $L_1$ \cite{Xu2010,XU2012}. $L_{0.5}$ regularization is still compatible with gradient descent (although it is no longer convex) and has been applied to neural networks \cite{Fan2014,Wu2014}. Experimental studies suggest that $L_{0.5}$ regularization performs no worse than other $L_q$ regularizers for $0<q<0.5$, so $L_{0.5}$ is optimal for sparsity \cite{XU2012}. Our experiments with $L_{0.3}$ and $L_{0.7}$ regularizers show no significant overall improvement compared to the $L_{0.5}$ regularizer, in agreement with this study. In addition, our experiments show that $L_{0.5}$ drives the solution towards sparsity more strongly than $L_1$ and produces much simpler expressions.

In particular, we use a smoothed $L_{0.5}$ proposed in \cite{Wu2014}, and we label their approach as $L_{0.5}^*$. The original $L_{0.5}$ regularization has a singularity in the gradient as the weights go to 0, which can make training difficult for gradient descent-based methods. To avoid this, the $L_{0.5}^*$ regularizer uses a piecewise function to smooth out the function at small magnitudes:

\begin{equation}
L_{0.5}^*(w)=
\begin{cases} 
      \lvert w \rvert^{1/2} & \lvert w \rvert \geq a \\
      \left(-\frac{w^4}{8a^3} + \frac{3w^2}{4a} + \frac{3a}{8}\right)^{1/2} & \lvert w \rvert < a
   \end{cases}
\label{eq:l12}
\end{equation}

\begin{figure}[t]
\centering
    \includegraphics[width=0.8\columnwidth]{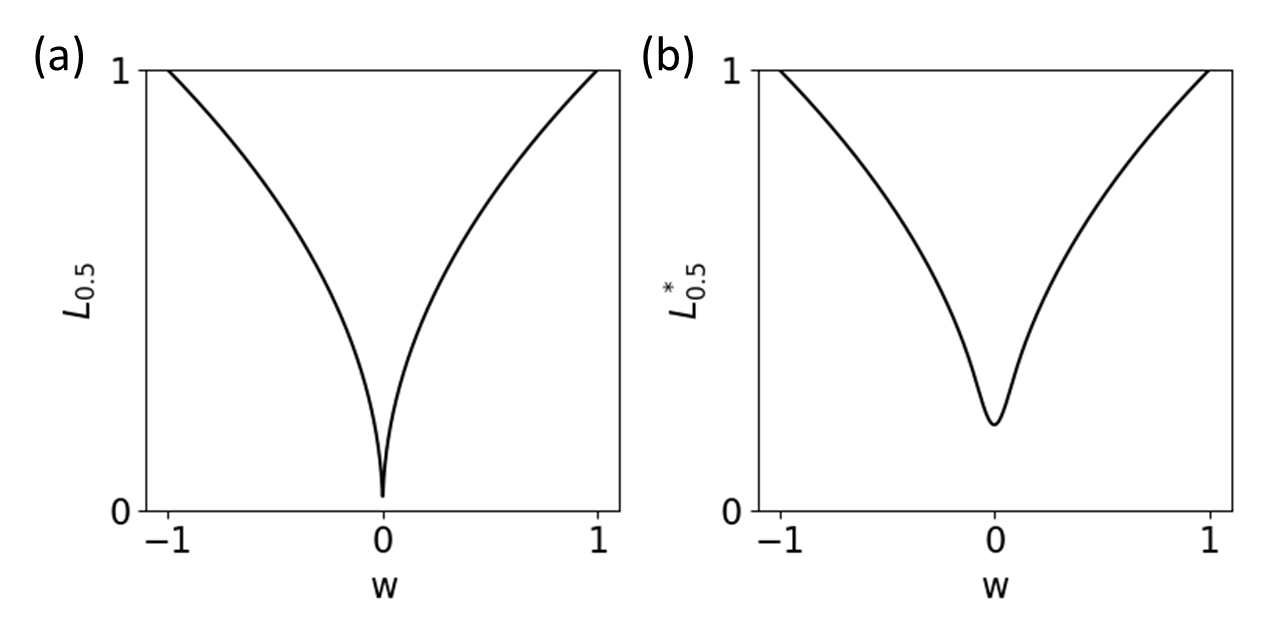}
	\caption{(a) $L_{0.5}$ and (b) $L_{0.5}^*$ regularization, as described in (\ref{eq:reg}) and (\ref{eq:l12}), respectively. The threshold for the plot of (\ref{eq:l12}) is set to $a=0.1$ for easy visualization, but we use a threshold of $a=0.01$ in our experiments.}
	\label{fig:reg}
\end{figure}

A plot of the $L_{0.5}$ and $L_{0.5}^*$ regularization are shown in Figure \ref{fig:reg}. The smoothed $L_{0.5}^*$ regularization avoids the extreme gradient values to improve training convergence. In our experiments, we set $a=0.01$. When the EQL network is integrated with other deep learning architectures, the regularization is only applied to the weights of the EQL network. 

We have also implemented an EQL network with the relaxed $L_0$ regularization proposed by \cite{Louizos2017}, the details of which can be found in Appendix \ref{app:l0}.

\section{Experiments}
\label{section:experiments}
\subsection{Symbolic Regression}
\label{sec:sr}

To validate the EQL network's ability to perform symbolic regression, we first test the EQL network on data generated by analytical functions such as $\exp{\left(-x^2\right)}$ or $x_1^2+\sin{\left(2\pi x_2\right)}$. The data is generated on the domain $x_i\in[-1, 1]$. Because of the network's sensitivity to random initialization of the weights, we run 20 trials for each function. 
We then count the number of times the network has converged to the correct answer ignoring small terms and slight variations in the coefficients from the true value. Additionally, equivalent answers (such as $\sin(4\pi + x)$ instead of $\sin(2\pi + x)$) are counted as correct. These results are shown in Appendix \ref{app:benchmark}.

The network only needs to be able to find the correct answer at least once over a reasonable number of trials, as one can construct a system that picks out the desired equation from the different trials by a combination of equation simplicity and generalization ability. The generalization ability is measured by the equation error evaluated on the domain $x_i\in[-2,2]$. This extrapolation error of the correct equation tends to be orders of magnitude lower than that of other equations that the network may find, making it simple to pick out the correct answer.

The network is still able to find the correct answer when 10\% noise is added to the data. We also test an EQL network with 3 hidden layers which still finds the correct expression and is able to find even more complicated expressions such as $\left(x_1+x_2x_3\right)^3$.

\subsection{MNIST Arithmetic}
\label{sec:mnist}

In the first experiment, we demonstrate the ability to combine symbolic regression and image recongition through an arithmetic task on MNIST digits. MNIST, a popular dataset for image recognition, can be notated as $\dataset=\{ \chi , \psi\}$, where $\chi$ are $28\times28$ greyscale images of handwritten digits and $\psi\in\{0,1,...,9\}$ is the integer-value label. Here, we wish to learn a simple arithmetic function, $y=\psi_1+\psi_2$, with the corresponding images $\{\chi_1,\chi_2\}$ as inputs, and train the system end-to-end such that the system learns how to ``add" two images together.

\begin{figure}[t]
\centering
    \includegraphics[width=\columnwidth]{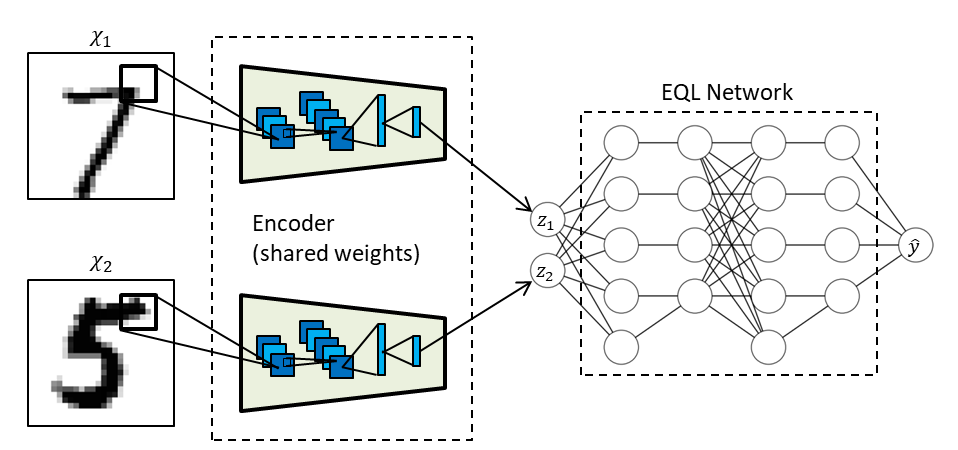}
	\caption{Schematic of the MNIST addition architecture. An encoder consisting of convolutional layers and fully-connected layers operate on each MNIST image and extract a single-dimensional latent variable. The two encoders share the same weights. The two latent variables are then fed into the EQL network. The entire system is fed end-to-end and without pre-training.}
	\label{fig:mnist}
\end{figure}

The deep learning architecture is shown in Figure \ref{fig:mnist}. The input to the system consists of two MNIST digits, $x=\{\chi_1,\chi_2\}$. During training, $\chi_i$ is randomly drawn from the MNIST training dataset. Each of $\{\chi_1,\chi_2\}$ are fed separately into an encoder to produce single-dimensional latent variables $\{z_1,z_2\}$ that are not constrained and can take on any real value, $z_{1,2} \in \mathbb{R}$. Alternatively, one can think of the system as having a separate encoder for each digit, where the two encoders share the same weights, as illustrated in Figure \ref{fig:mnist}. The encoder consists of two convolutional layers with max pooling layers followed by two fully-connected layers and a batch normalization layer at the output. More details on the encoder can be found in Appendix \ref{app:data}. The latent variables $\{z_1, z_2\}$ are then fed as inputs into the EQL network. The EQL network has a single scalar output $\hat{y}$ which is compared to the true label $y=\psi_1+\psi_2$.

The entire network is trained end-to-end using a mean-squared error loss between the predicted label $\hat{y}$ and the true label $y$. In other words, the encoder is not trained separately from the EQL network. Note that the encoder closely resembles a simple convolutional neural network used for classifying MNIST digits except that it outputs a scalar value instead of logits that encode the digit. Additionally, there is no constraint on the properties of $z_{1,2}$, but we expect that it has a one-to-one mapping to the true label $\psi_{1,2}$.

\subsection{Dynamical System Analysis}

\begin{figure}[htbp]
	\centering
    \includegraphics[width=\columnwidth]{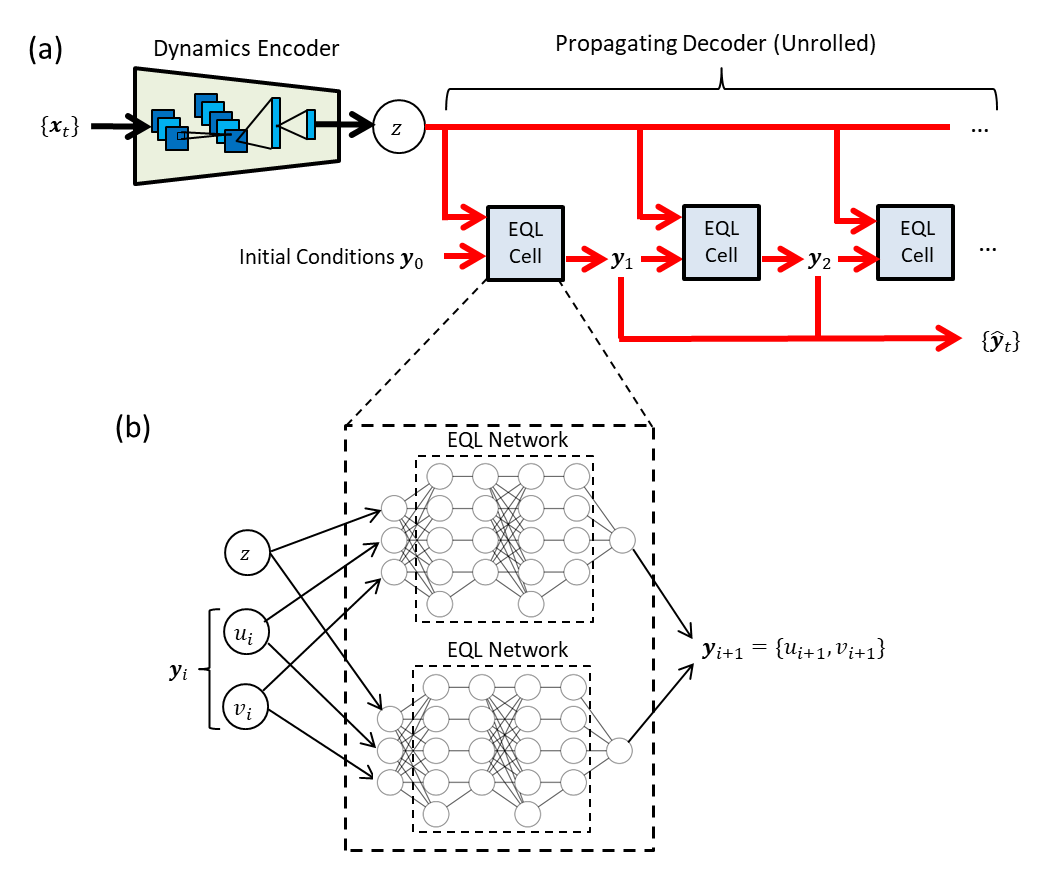}
	\caption{(a) Architecture to learn the equations that propagate a dynamical system. (b) Each EQL cell in the propagating decoder consists of a separate EQL network for each dimension of $\mathbf{y}$ to be predicted. In our case, $\mathbf{y}=\{u,v\}$ where $u$ is the position and $v$ is velocity, so there are 2 EQL networks in each EQL cell.}
	\label{fig:depd}
\end{figure}

A potentially powerful application of deep learning in science exploration and discovery is discovering parameters in a dynamical system in an unsupervised way and using these parameters to predict the propagation of a similar system. For example, \cite{Zheng2018} uses multilayer perceptrons to extract relevant properties from a system of bouncing balls (such as the mass of the balls or the spring constant of a force between the balls) and simultaneously predict the trajectory of a different set of objects. 
\cite{Lu2019} accomplishes a similar goal but using a dynamics encoder (DE) with convolutional layers and a propagating decoder (PD) with deconvolutional layers to enable analysis and prediction of spatiotemporal systems such as those governed by PDEs.
This architecture is designed to analyze spatiotemporal systems that may have an uncontrolled dynamical parameter that varies among different instances of the dataset such as the diffusion constant in the diffusion equation. The parameters encoded in a latent variable are fed into the PD along with a set of initial conditions, which the PD propagates forward in time based on the extracted physical parameter and learned dynamics.

Here, we present a deep learning architecture shown in Figure \ref{fig:depd} which is based on the DE-PD architecture. 
The DE takes in the full input series $\{\mathbf{x}_t\}_{t=0}^{T_x}$ and outputs a single-dimensional latent variable $z$. Unlike the original DE-PD architecture presented in \cite{Lu2019}, the DE here is not a VAE. The DE here consists of several convolutional layers followed by fully-connected layers and a batch normalization layer. More details are given in Appendix \ref{app:data}. 
The parameter $z$ and an initial condition $\mathbf{y}_0$ are fed into the PD which predicts the future time steps $\{\hat{\mathbf{y}}_t\}_{t=1}^{T_y}$ based on the learned dynamics. 
The PD consists of a ``EQL cell" in a recurrent structure, such that each step in the recurrent structure predicts a single time step forward. The EQL cell consists of separate EQL networks for each of feature, or dimension, in $\hat{\mathbf{y}}_t$.

The full architecture is trained end-to-end using a mean-squared error loss between the predicted dynamics $\{\hat{\mathbf{y}}_t\}_{t=1}^{T_y}$ and the target series $\{\mathbf{y}_t\}_{t=1}^{T_y}$. Similar to the architecture in Section \ref{sec:mnist}, the DE and PD are not trained separately, and there is no additionaly restriction or bias placed on the latent variable $z$. The datasets are derived from two different physical systems (kinematics and simple harmonic oscillator) as described in the following sections.

\begin{figure}[ht]
\centering
    \includegraphics[width=\columnwidth]{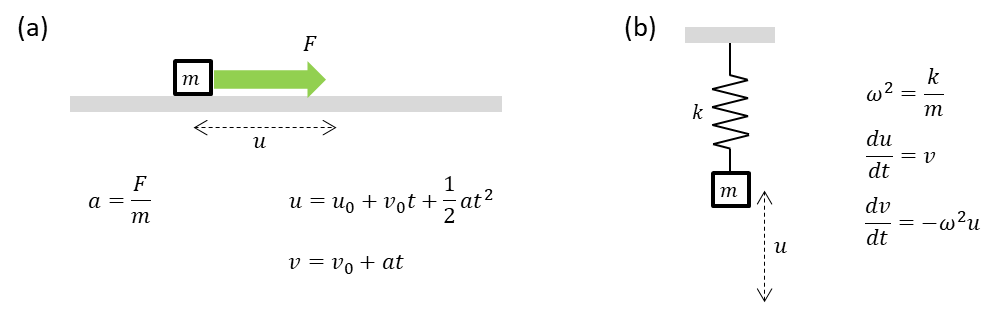}
	\caption{(a) Kinematics describes the dynamics of an object where a force $F$ is applied to a mass $m$. (b) Simple harmonic oscillator describes a mass $m$ on a spring with spring constant $k$. In both cases, $u$ is the displacement of the mass and $v$ is the velocity.}
	\label{fig:physics}
\end{figure}

\subsubsection{Kinematics}

Kinematics describes the motion of objects and is used in physics to study how objects move under an applied force. A schematic of a physical scenario described by kinematics is shown in Figure \ref{fig:physics}(a) in which an object on a frictionless surface has a force applied to it where the direction of the force is parallel to the surface. The relevant parameter to describe the object's motion can be reduced to $a=\frac{F}{m}$ for a constant force $F$ and object mass $m$. Given position $u_i$ and velocity $v_i$ at time step $i$, the object's state at time step $i+1$ are given by
\begin{equation}
\begin{aligned}
u_{i+1} &= u_i+v_i \Delta t+\frac{1}{2}\Delta t^2 \\
v_{i+1} &= v_i + a \Delta t 
\label{eq:kinematics}
\end{aligned}
\end{equation}

Acceleration $a$ varies across different instances of the dataset. In our simulated dataset, we draw initial state and acceleration from uniform distributions:
\[u_0, v_0, a \sim \mathcal{U}(-1,1) \]
We set $\Delta t=1$. The initial parameters $u_0, v_0$ are fed into the propagator, and the dynamics encoder output is expected to correlate with $a$. 

\subsubsection{Simple Harmonic Oscillator (SHO)}

The second physical system we analyze to demonstrate the dynamic system analysis architecture is the simple harmonic oscillator (SHO), a ubiquitous model in physics that can describe a wide range of physical systems including springs, pendulums, particles in a potential, or electric circuits. In general, the dynamics of the SHO can be given by the coupled first-order ordinary differential equation (ODE)
\begin{equation}
\begin{aligned}
\frac{du}{dt} &= v \\
\frac{dv}{dt} &= -\omega^2 u
\label{eq:sho-ode}
\end{aligned}
\end{equation}
\noindent where $u$ is the position, $v$ is the velocity, and $\omega$ is the resonant frequency of the system. In the case of a spring as shown in Figure \ref{fig:physics}(b), $\omega=\sqrt{k/m}$ where $k$ is the spring constant and $m$ is the mass of the object on the end of the spring. 

The SHO system can be solved for numerically using a finite-difference approximation for the time derivatives. For example, the Euler method for integrating Eqs. \ref{eq:sho-ode} gives:
\begin{equation}
\begin{aligned}
u_{i+1} &= u_i + v\Delta t \\
v_{i+1} &= v_i - \omega^2 u \Delta t
\end{aligned}
\end{equation}

In our experiments, we generate data with parameters drawn from uniform distributions:
\begin{align*}
u_0, v_0 & \sim \mathcal{U}(-1,1) \\ 
\omega^2 & \sim \mathcal{U}(0.1,1)
\end{align*}
The state variables $u$ and $v$ are measured at a time step of $\Delta t=0.1$ to allow the system to find the finite-difference solution. Because of this small time step, we also need to propagate the solution for more time steps to find the right equation (otherwise the system learns the identity function). To avoid problems of the recurrent structure predicting a solution that explodes oward $\pm \infty$, we start the training with propagating only 1 time step, and add more time steps as the training continues. This is a similar strategy as \cite{Long2018-2} except that we are not restarting the training.

The initial parameters $u_0, v_0$ are fed into the propagator, and the dynamics encoder output is expected to correlate with $\omega^2$.

\subsection{Training}

The neural network is implemented in TensorFlow \cite{tensorflow2015-whitepaper}. The network is trained using backpropagation with the RMSProp optimizer \cite{tieleman2012lecture} and the following loss function:
\[\mathcal{L}=\frac{1}{N}\sum\left(y_i-\hat{y}_i\right)^2 + \lambda L_{0.5}^* \]
\noindent where $N$ is the size of the training dataset and $\lambda$ is a hyperparameter that balances the regularization versus the mean-squared error.

Similar to \cite{Martius2016}, we introduce a multi-phase training schedule. In an optional first phase, we train with a small value of $\lambda$, allowing for the parts of the network apart from the EQL to evolve freely and extract the latent parameters during training. In the second phase, $\lambda$ is increased to a point where it forces the EQL network to become sparse. After this second phase, weights in the EQL network below a certain threshold $\alpha$ are set to 0 and frozen such that they stay 0, equivalent to fixing the $L_0$ norm. In the final phase of training, the system continues training without $L_{0.5}^*$ regularization (i.e. $\lambda=0$) and with a reduced maximum learning rate in order to fine-tune the weights.  

Specific details for each experiment are listed in Appendix \ref{app:data}.

\section{Results}

\subsection{MNIST Arithmetic}
\label{sec:mnist_result}

\begin{figure}[t]
\centering
    \includegraphics[width=0.8\columnwidth,trim=3 0 0 2,clip]{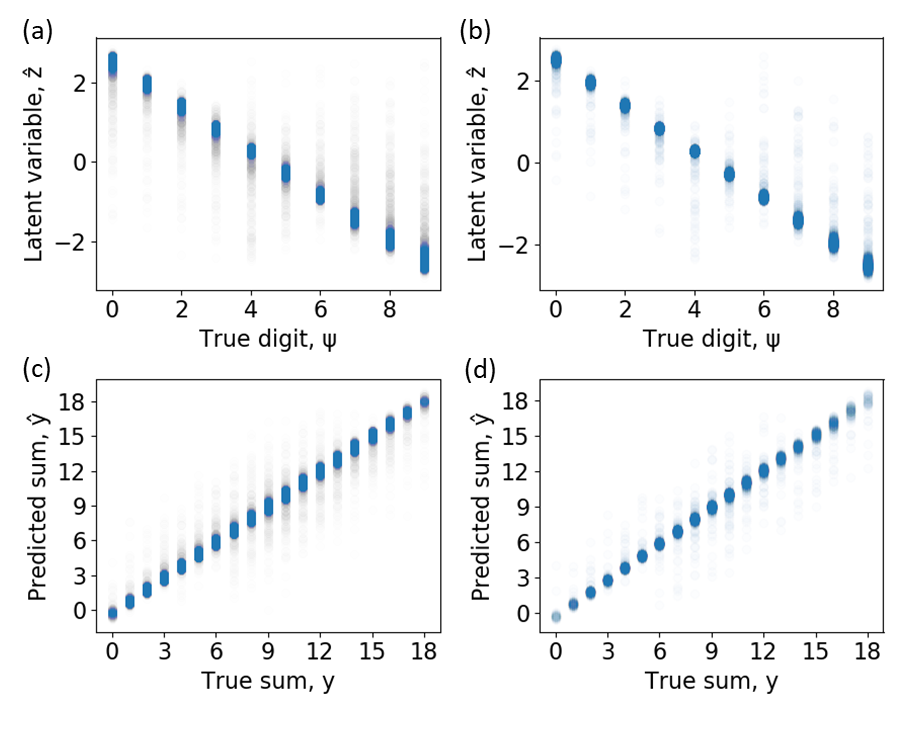}
	\caption{The ability of the encoder to differentiate between digits as measured by the latent variable $z$ versus the true digit $\psi$ for digits $\chi$ drawn from the MNIST (a) training dataset and (b) test dataset. The correlation coefficients are $-0.985$ and $-0.988$, respectively. The ability of the entire architecture to fit the label $y$ as measured by the predicted sum $\hat{y}$ versus the true sum $y$ for digits $\chi$ drawn from the MNIST (c) training dataset and (d) MNIST test dataset.}
	\label{fig:mnist_result}
\end{figure}

Figure \ref{fig:mnist_result}(b) plots the latent variable $z$ versus the true label $\psi$ for each digit after the entire network has been trained. Note that the system is trained on digits drawn from the MNIST training dataset and we also evaluate the trained network's performance on digits drawn from the MNIST test dataset. We see a strong linear correlation for both datasets, showing that the encoder has successfully learned a linear mapping to the digits despite not having access to the digit label $\psi$. Also note that there is a constant scaling factor between $z$ and $\psi$ due to the lack of constraint on $z$. A simple linear regression shows that the relation is 
\begin{equation}
\psi=-1.788z+4.519
\label{eq:mnist}
\end{equation}

\begin{table}[htbp]
\centering
\caption{MNIST Arithmetic Expected and Extracted Equations}
\begin{tabular}{ll}
\toprule
True & $y=\psi_1 + \psi_2$ \\ \hline
Encoder   & $\hat{y} = -1.788 z_1 - 1.788 z_2 + 9.04$ \\ \hline
EQL		& 	$\hat{y}=-1.809 z_1 - 1.802 z_2 + 9$ \\
\bottomrule
\end{tabular}
\label{table:mnist}
\end{table}

The extracted equation from the EQL network for this result is shown in Table \ref{table:mnist}. The ``Encoder" equation is what we expect based on the encoder result in Equation \ref{eq:mnist}. 
We conclude that the EQL network has successfully extracted the additive nature of the function. Plotted in Figure \ref{fig:mnist_result}(c-d) are the predicted sums $\hat{y}$ versus the true sums $y$. The mean absolute errors of prediction for the system drawing digits from the MNIST training and test datasets are 0.307 and 0.315, respectively.

While the architecture is trained as a regression problem using a mean square loss, we can still report accuracies as if it is a classification task since the labels $y$ are integers. To calculate accuracy, we first round the predicted sum $\hat{y}$ to the nearest integer and then compare it to the label $y$. The trained system achieves accuracies of 89.7\% and 90.2\% for digits drawn from the MNIST training and test datasets, respectively.

To demonstrate the generalization of this architecture to data outside of the training dataset, we train the system using a scheme where MNIST digit pairs ${\chi_1,\chi_2}$ are randomly sampled from the MNIST training dataset and used as a training data point if they follow the condition $\psi_1+\psi_2<15$. Otherwise, the pair is discarded. In the test phase, MNIST digit pairs ${\chi_1,\chi_2}$ are randomly sampled from the MNIST training dataset and kept in the evaluation dataset if $\psi_1+\psi_2\geq15$. Otherwise, the pair is discarded.

For comparison, we also test the generalization of the encoder by following the above procedures but drawing MNIST digit pairs ${\chi_1,\chi_2}$ from the MNIST test dataset.

\begin{table}[ht]
\centering
\caption{MNIST Arithmetic Generalization Results}
\begin{tabular}{llSS}\toprule
\multicolumn{2}{c}{} & \multicolumn{2}{c}{Accuracy [\%]}\\
\cmidrule{3-4}
\begin{tabular}[c]{@{}l@{}}Source of $w_i$ to form\\ $x=\{w_1,w_2\}$\end{tabular} & \begin{tabular}[c]{@{}l@{}}Network after \\ the encoder\end{tabular} & {$y<15$} & {$y\geq15$} \\ \midrule
\multirow{2}{*}{MNIST training dataset}                            & EQL                                                                                 & 92                            & 87                        \\ 
                                                                   & ReLU                                                                                 & 93                            & 0.8                         \\ 
\multirow{2}{*}{MNIST test dataset}                                & EQL                                                                                 & 91                            & 83                       \\ 
                                                                   & ReLU                                                                               & 92                            & 0.6                        \\ \bottomrule
\end{tabular}
\label{table:mnist_general}
\end{table}

Generalization results of the network are shown in Table \ref{table:mnist_general}. In this case, the EQL network has learned the equation $\hat{y}=-1.56z_1-1.56z_2+8.66$.
First, the most significant result is the difference between the accuracy evaluated on pairs $y<15$ and pairs $y\geq15$. For the architecture with the EQL network, the accuracy drops by a few percentage points. However, for the architecture where the EQL network is replaced by the commonly used fully-connected network with ReLU activation functions (which we label as ``ReLU"), the accuracy drops to below $1\%$ showing that the results of the EQL is able to generalize reasonably well in a regime where the ReLU cannot generalize at all. It is not necessarily an issue with the encoder since the system sees all digits 0 through 9.

Second, the accuracy drops slightly when digits are drawn from the MNIST test dataset versus when the digits are drawn from the MNIST training dataset, as expected. We did not optimize the hyperparameters of the digit extraction network since the drop in accuracy is small. Therefore, this could be optimized further if needed.

Finally, the accuracy drops slightly for pairs $y<15$ when using the EQL versus the ReLU network. This is unsurprising since the larger size and symmetric activation functions of the ReLU network constrains the network less than the EQL and may make the optimization landscape smoother.

\subsection{Kinematics}

\begin{figure}[htbp]
\centering
    \includegraphics[width=\columnwidth, trim=3 0 3 3,clip]{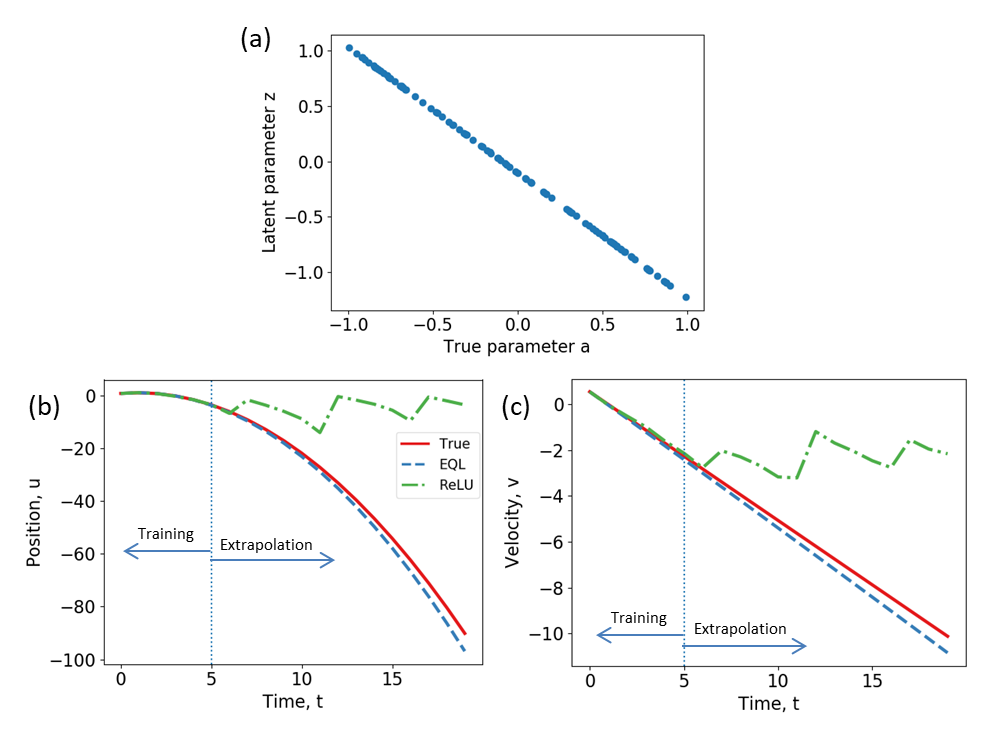}
	\caption{(a) Latent parameter $z$ of the dynamic encoder architecture after training plotted as a function of the true parameter $a$. We see a strong linear correlation. (b,c) Predicted propagation $\{\hat{\mathbf{y}}_i\}=\{\hat{u}_i,\hat{v}_i\}$ with the EQL cell and a conventional network using ReLU activations. ``True" refers to the true propagation $\{\mathbf{y}_i\}$.}
	\label{fig:kinematics}
\end{figure}

Figure \ref{fig:kinematics}(a) shows the extracted latent parameter $z$ plotted as a function of the true parameter $a$. We see a linear correlation with correlation coefficient close to $-1$, showing that the dynamics encoder has extracted the relevant parameter of the system. Again, there is a scaling relation between $z$ and $a$:
\begin{equation}
a=-0.884z-0.091
\label{eq:kinematics-scale}
\end{equation}

An example of the equations found by the EQL network after training is shown in Table \ref{table:kinematics}. The ``DE" equation is what we expect based on latent variable extracted with the relation in Equation \ref{eq:kinematics-scale}. These results match closely with what we expect.

\begin{table}[htbp]
\centering
\caption{Kinematics Expected and Extracted Equations}
\begin{tabular}{ll}
\toprule
True & \begin{tabular}[c]{@{}l@{}} $u_{i+1} =u_i+v_i+\frac{1}{2}a$ \\
$v_{i+1} = v_i + a$ \end{tabular} \\ \hline
DE   & \begin{tabular}[c]{@{}l@{}}$\hat{u}_{i+1} = u_i + v_i - 0.442z - 0.045$ \\ $\hat{v}_{i+1} = v_i -0.884z-0.091$\end{tabular}     \\ \hline
EQL   & \begin{tabular}[c]{@{}l@{}} $\hat{u}_{i+1} = 1.002u_i + 1.002v_i - 0.475z$ \\ $\hat{v}_{i+1} = 1.002v_i - 0.918z - 0.102$ \end{tabular}     \\ \bottomrule
\end{tabular}
\label{table:kinematics}
\end{table}

The predicted propagation $\{\hat{\mathbf{y}}_i\}$ is plotted in Figure \ref{fig:kinematics}(c-d). ``True" is the true solution that we want to fit, and ``EQL" is the solution propagated by the EQL network. For comparison, we also train a neural network with a similar architecture to the one shown in Figure \ref{fig:depd} but where the EQL cell is replaced by a standard fully-connected neural network with 2 hidden layers of 50 neurons each and ReLU activation functions (which we label as ``ReLU"). While both networks match the true solution very closely in the training regime (left of the dotted line), the ReLU network quickly diverges from the true solution outside of the training regime. The EQL cell is able to match the solution reasonably well for several more time steps, showing how it can extrapolate beyond the training data.

\subsection{SHO}

\begin{figure}[htbp]
\centering
    \includegraphics[width=\columnwidth,trim=1 0 0 0,clip]{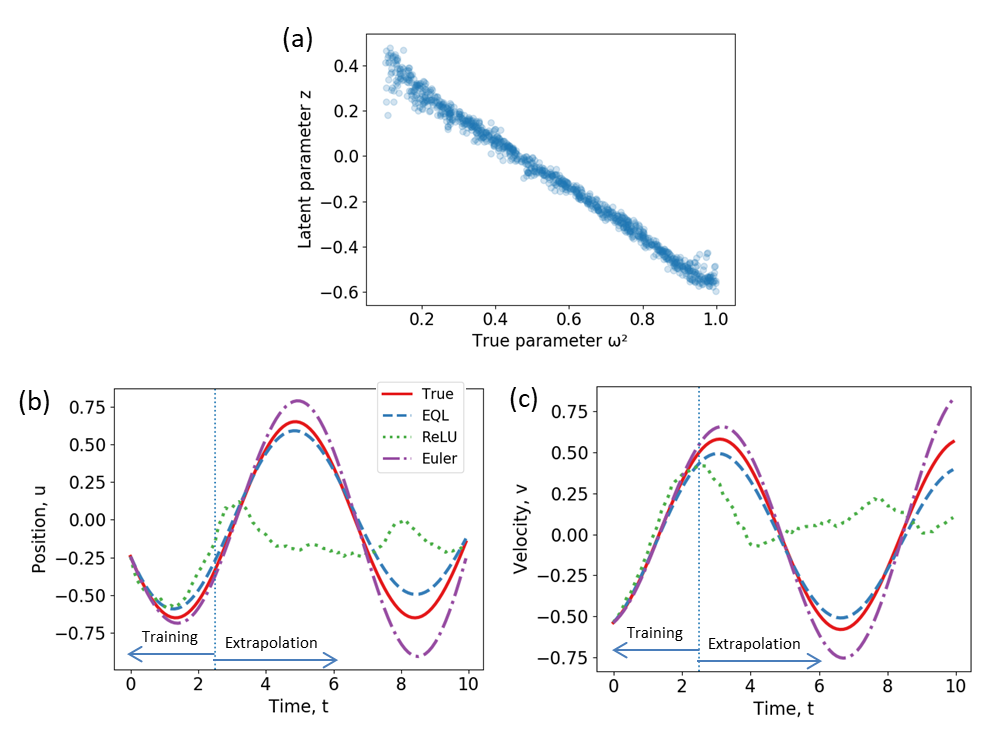}
	\caption{Results of training on the SHO system. (a) Latent parameter $z$ of the dynamic encoder architecture after training plotted as a function of the true parameter $\omega^2$. We see a good linear correlation. (b) Position $u$ and (c) velocity $v$ as a function of time for various models. ``True" refers to the analytical solution. ``EQL" refers to the propagation equation discovered by the EQL network. ``ReLU" refers to propagation by a conventional neural network that uses ReLU activation functions. ``Euler" refers to finite-difference solution using the Euler method. }
	\label{fig:sho}
\end{figure}

The plot of the latent variable $z$ as a function of the true parameter $\omega^2$ is shown in Figure \ref{fig:sho}(a). Note that there is a strong linear correlation between $z$ and $\omega^2$ as opposed to between $z$ and $\omega$. This reflects the fact that using $\omega^2$ requires fewer operations in the propagating equations than $\omega$, the latter of which would require a squaring function. Additionally, the system was able to find that $\omega^2$ is the simplest parameter to describe the system due to the sparsity regularization on the EQL cell. We see a strong linear correlation with a correlation coefficient of $-0.995$, showing that the dynamics encoder has successfully extracted the relevant parameter of the SHO system. A linear regression shows that the relation is:
\begin{equation}
\omega^2=-0.927z+0.464
\label{eq:sho}
\end{equation}

\begin{table}[htbp]
\centering
\caption{SHO Expected and Extracted Equations}
\begin{tabular}{ll}
\toprule
True & 
$\begin{aligned}
u_{i+1} &= u_i + 0.1v_i \\
v_{i+1} &= v_i - 0.1 \omega^2 u_i
\end{aligned}$ \\ \midrule
DE   & 
$\begin{aligned}
\hat{u}_{i+1} &=  u_i + 0.1v_i \\ 
\hat{v}_{i+1} &=  v_i - 0.0464 u_i + 0.0927 u_i z     
\end{aligned}$
\\ \midrule
DE, 2nd Order   & 
$\begin{aligned}
\hat{u}_{i+1} &=  u_i + 0.1v_i \\ 
\hat{v}_{i+1} &=  0.998v_i - 0.0464 u_i  + 0.0927 u_i z \\
 	& \qquad {}  + 0.0046 v_iz
\end{aligned}$ \\ \midrule
EQL    &  
$\begin{aligned}
\hat{u}_{i+1} &=  0.994u_i+0.0992v_i-0.0031 \\ 
\hat{v}_{i+1} &=  0.995v_i-0.0492d+0.084 u_i z  \\
	& \qquad {} + 0.0037v_iz + 0.0133 z^2
\end{aligned}$      \\ \bottomrule
\end{tabular}
\label{table:sho}
\end{table}

The equations extracted by the EQL cell (consisting of 2 EQL networks) are shown in Table \ref{table:sho}. The ``DE" equation is what we expect based on the dynamics encoder result in Equation \ref{eq:sho}. Immediately, we see that the expression for $\hat{u}_{i+1}$ and the first three terms of $\hat{v}_{i+1}$ match closely with the Euler method approximation using the latent variable relation extracted by the dynamics encoder.

An interesting point is that while we normally use the first-order approximation of the Euler method for integrating ODEs:
\[v_{i+1}=v_i+\Delta t \left.\frac{dv}{dt}\right|_{t=i} + \mathcal{O}(\Delta t^2) \]
\noindent it is possible to expand the approximation to find higher-order terms. If we expand the Euler method to its second-order approximation, we get:
\begin{align*}
v_{i+1} &= v_{i}+\Delta t \left.\frac{dv}{dt}\right|_{t=i} + \frac{1}{2}\Delta t^2 \left.\frac{d^2v}{dt^2}\right|_{t=i} +  \mathcal{O}(\Delta t^3 ) \\
&\approx v_i - \Delta t \omega^2 u_i - \frac{1}{2} \Delta t^2 \omega^2 v_i 
\end{align*}

The expected equation based on the dynamics encoder result and assuming the 2nd order expansion is labeled as ``DE, 2nd Order" in Table \ref{table:sho}. It appears that the EQL network in this case has not only found the first-order Euler finite-difference method, it has also added on another small term that corresponds to second-order term in the Taylor expansion of $v_{i+1}$. The last term found by the EQL network, $0.0133z^2$ is likely from either cross-terms inside the network or a lack of convergence to exactly 0 and would likely disappear with another thresholding process.

The solution propagated through time is shown in Figure \ref{fig:sho}(b-c). As before, ``ReLU" is the solution propagated by an architecture where the EQL network is replaced by a conventional neural network with 4 hidden layers of 50 units each and ReLU activation functions. For an additional comparison, we have also calculated the finite-difference solution using Euler's method to integrate the true ODEs which is labeled as ``Euler". 

Within the training regime, all of the methods fit the true solution reasonably well. However, the conventional neural network  with ReLU activation functions completely fails to extrapolate beyond the training regime and essentially regresses to noise around 0. The Euler method and the EQL network are both able to extrapolate reasonably well beyond the training regime, although they both start to diverge from the true solution due to the large time step and the accumulated errors. A more accurate method such as the Runge-Kutta method almost exactly fits the analytical solution, which is not surprising due to its small error bound. However, it is more complex than the Euler method and would likely require a larger EQL network to find an expression similar to the Runge-Kutta method. Interestingly, the EQL network solution has a smaller error than the Euler solution, demonstrating that the EQL network was able to learn higher-order corrections to the first-order Euler method. This could possibly lead to discovery of more efficient integration schemes for differential equations that are difficult to solve through finite-difference. 

\section{Conclusion}
We have shown how we can integrate symbolic regression with deep learning architectures and train the entire system end-to-end to take advantage of the powerful deep learning training techniques that have been developed in recent years. Namely, we show that we can learn arithmetic on MNIST digits where the system must learn to identify the images in an image recognition task while simultanesouly extracting the mathematical expression that relates the digits to the answer. Additionally, we show that we can simultaneously extract an unknown parameter from a dynamical system extract the propagation equations. In the SHO system, the results suggest that we can discover new techniques for integrating ODEs, potentially paving the way for improved integrators, such as integrators for stiff ODEs that may be difficult to solve with numerical methods.

One direction for future work is to study the role of random initializations and make the system less sensitive to random initializations. As seen by the benchmark results of the EQL network in Appendix \ref{app:benchmark}, the EQL network is not always able to find the correct mathematical expression. This is because there are a number of local minima in the EQL network that the network can get stuck in, and gradient-based optimization methods are only guaranteed to find local minima rather than global minima. Local minima are not typically a concern for neural networks because the local minima are typically close enough in performance to the global minimum \cite{Choromanska2015}. However, for the EQL network, we often want to find the true global minimum. In this work, we have alleviated this issue by increasing stochasticity through large learning rates and by decreasing the sensitivity to random initializations by duplicating activation functions. Additionally, we run multiple trials and find the best results, either manually or through an automated system \cite{Martius2016,Sahoo2018}. In future work, it may be possible to find the true global minimum without resorting to multiple trials as it has been shown that over-parameterized neural networks with certain types of activation functions are able to reach the global minimum through gradient descent in linear time regardless of the random initialization \cite{Du2018}.

Other directions for future work include expanding the types of deep learning architectures that the EQL network can integrate with. For example, supporting spatio-temporal systems can lead to PDE discovery. The spatial derivatives could be calculated using known finite-difference approximations or learnable kernels \cite{Long2018}. Another possible extension is to introduce parametric dependence in which unknown parameters have a time-dependence, which has been studied in PDE-discovery using group sparsity \cite{Rudy2019}. Additionally, the encoder can be expanded to capture a wider variety of data such as videos \cite{Chari2019}, audio signals, and text.

\appendices
\section{EQL Network Details}
\label{app:benchmark}

The activation functions in each hidden layer consist of:
\begin{multline*} 
[1 (\times 2), g (\times 4), g^2 (\times 4), \sin (2\pi g) (\times 2), \\
e^g (\times 2), \text{sigmoid}(20g) (\times 2) , g_1*g_2 (\times 2)] 
\end{multline*}
\noindent where the sigmoid function is defined as:
\[\text{sigmoid}(g)=\frac{1}{1+e^{-g}}\]
\noindent and the $(\times i)$ indicated the number of times each activation function is duplicated. The $\sin$ and $\text{sigmoid}$ functions have multipliers inside so that the functions more accurately represent their respective shapes inside the input domain of $x\in[-1,1]$. Unless otherwise stated, these are the activation functions used for the other experiments as well. The exact number of duplications is arbitrary and does not have a significant impact on the system's performance. Future work may include experimenting with a larger number of duplications.

We use two phases of training, where the first phase has a learning rate of $10^{-2}$ and regularization weight $5\times10^{-3}$ for 2000 iterations. Small weights are frozen and set to 0 after the first phase. The second phase of training has a learning rate of $10^{-3}$ for 10000 iterations.

To benchmark our symbolic regression system, we choose a range of trial functions that our architecture can feasibly construct, train the network through 20 trials, and count how many times it reaches the correct answer. Benchmarking results are shown in Table \ref{tab:benchmark}. As mentioned in section \ref{sec:sr}, we only need the network to find the correct equation at least once since we can construct a system that automatically picks out the correct solution based on equation simplicity and test error.

\begin{table}[htbp]
\centering
\caption{Benchmark results for the EQL network.}
\begin{tabular}{lSS}
\toprule
& \multicolumn{2}{c}{Success Rate}\\
\cmidrule{2-3}
Function &  {$L_{0.5}$} & {$L_0$} \\ \midrule
$x$							& 1     & 1\\ 
$x^2$						& 0.6   & 0.75\\ 
$x^3$						& 0.3   & 0.05\\ 
$\sin(2\pi x)$				& 0.45  & 0.85\\ 
$xy$							& 0.8   & 1\\ 
$\frac{1}{1+e^{-10x}}$		& 0.3   & 0.55\\ 
$\frac{xy}{2}+\frac{z}{2}$	& 0.05  & 0.95\\ 
$\exp(-x^2)$					& 0.05  & 0.15\\ 
$x^2+\sin(2\pi y)$			& 0.2   & 0.8\\ 
$x^2+y-2z$					& 0.6   & 0.9\\ \bottomrule
\end{tabular}
\label{tab:benchmark}
\end{table}

\subsection{Computational Efficiency}

With respect to the task of symbolic regression, we should note that this algorithm does not offer an asymptotic speedup over conventional symbolic regression algorithms, as the problem of finding the correct expression requires a combinatorial search over the space of possible expressions and is NP-hard. Rather, the advantage here is that by solving symbolic regression problems through gradient descent, we can integrate symbolic regression with deep learning architectures.

Experiments are run on an Nvidia GTX 1080 Ti. Training the EQL network with 2 hidden layers ($L=2$) for 20,000 epochs takes 37 seconds, and training the EQL network with 3 hidden layers ($L=3$) takes 51 seconds. 

In general, the computational complexity of the EQL network itself is the same as that of a conventional fully-connected neural network. The only difference are the activation functions which are applied by iterating over $\mathbf{g}$ and thus takes $\mathcal{O}(n)$ time where $n$ is the number of nodes in each layer. However, the computational complexity of a neural network is dominated by the weight matrix multiplication which takes $\mathcal{O}(n^2)$ time for both the EQL network and the conventional fully-connected neural network.

\section{Relaxed $L_0$ Regularization}
\label{app:l0}

We have also implemented an EQL network that uses a relaxed form of $L_0$ regularization for neural networks introduced by \cite{Louizos2017}. We briefly review the details here, but refer the reader to \cite{Louizos2017} for more details.

The weights $\mathbf{W}$ of the neural network are reparameterized as
\[\mathbf{W}=\mathbf{\tilde{W}}\odot \mathbf{z}\]
\noindent where ideally each element of $\mathbf{z}$, $z_{j,k}$, is a binary ``gate", $z_{j,k}\in\{0,1\}$. However, this is not differentiable and so we allow $z_{j,k}$ to be a stochastic variable drawn from the hard concrete distribution:
\begin{gather*}
u\sim\mathcal{U}(0,1) \\
s=\text{sigmoid}\left(\left[\log u - \log (1-u) + \log \alpha_{j,k}\right] / \beta \right) \\
\bar{s}=s(\zeta-\gamma)+\gamma) \\
z_{j,k} = \min(1, \max(0, \bar{s}))
\end{gather*}
\noindent where $\alpha_{j,k}$ is a trainable variable that describes the location of the hard concrete distribution, and $\beta, \zeta, \gamma$ are hyperparameters that describe the distribution. In the case of binary gates, the regularization penalty would simply be the sum of $\mathbf{z}$ (i.e., the number of non-zero elements in $\mathbf{W}$. However, in the case of the hard concrete distribution, we can calculate an analytical form for the expectation of the regularization penalty over the distribution parameters. The total loss function is then
\[\mathcal{L}=\frac{1}{N}\sum\left(y_i-\hat{y}_i\right)^2 + \sum_{j,k} \text{sigmoid}\left(\log\alpha_{j,k} - \beta \log \frac{-\gamma}{\zeta} \right) \]

The advantage of $L_0$ regularization is that it enforces sparsity without placing a penalty on the magnitude of the weights. This also allows us to train the system without needing a final stage where small weights are set to 0 and frozen. While the reparameterization by \cite{Louizos2017} requires us to double the number of trainable parameters in the neural network, the regularization is only applied to the EQL network which is small compared to the rest of the architecture.

In our experiments, we use the hyperparameters for the $L_0$ regularization suggested by \cite{Louizos2017}, although these can be optimized in future work. Additionally, while \cite{Louizos2017} apply group sparsity to the rows of the weight matrices with the goal of computational efficiency, we apply parameter sparsity with the goal fo simplifying the symbolic expression. We benchmark the EQL network using $L_0$ regularization with the aforementioned trial functions and list the results in Table \ref{tab:benchmark}. The success rates appear to be as good or better than the network using $L_{0.5}$ regularization for most of the trial functions that we have picked. We have also integrated the EQL network using $L_0$ regularization into the MNIST arithmetic and kinematics architectures, and have found similar results as the EQL network using $L_{0.5}$ regularization.

\section{Experiment Details}
\label{app:data}

\subsection{MNIST Arithmetic}

The encoder network consists of a convolutional layer with 32 $5\times5$ filters followed by a max pooling layer, another convolutional layer with 64 $5\times5$ filters followed by a max pooling layer, and 2 fully-connected layers with 128 and 16 units each with ReLU activation units. The max pooling layers have pool size of 2 and stride length of 2. The fully-connected layers are followed by 1-unit layer with batch normalization. The output of the batch normalization layer is divided by 2 such that the standard deviation of the output is 0.5. This decreases the range of the inputs to the EQL network since the EQL network was constructed assuming an input domain of $x\in[-1, 1]$. Additionally, the output of the EQL network, $\hat{y}^*$, is scaled as $\hat{y}=9\hat{y}^*+9$ before being fed into the loss function so as the normalize the output against the range of expected $y$ (this is equivalent to normalizing $y$ to the range $[-1, 1]$).

The ReLU network that is trained in place of the EQL network for comparison consists of two hidden layers with 50 units each and ReLU activation.

We use two phases of training, where the first phases uses a learning rate of $10^{-2}$ and regularization weight $\lambda=0.05$. The second phase uses a learning rate of $10^{-4}$ and no regularization. The small weights are frozen between the first and second phase with a threshold of $\alpha=0.01$. Each phase is trained for 10000 iterations.

\subsection{Kinematics}

To generate the kinematics dataset, we sample 100 values for $a$ and generate a time series $\{\mathbf{x}_t\}_{t=0}^{T_x-1}$ and $\{\mathbf{y}_t\}_{t=0}^{T_y}$ for each $a$. The input series is propagated for $T_x=100$ time steps. 

The dynamics encoder consists of 2 1D convolutional layers with 16 filters of length 5 in each layer. These are followed by a hidden layer with 16 nodes and ReLU activation function, an output layer with one unit, and a batch normalization layer with standard deviation 0.5. The ReLU network that is trained in place of the EQL network for comparison is the same as that of the MNIST task.

We use two phases of training, where the first phase uses a learning rate of $10^{-2}$ and a regularization weight of $\lambda=10^{-3}$ for a total of 5000 iterations. The system is trained on $T_y=1$ time step for the first 1000 iterations, and then $T_y=5$ time steps for the remainder of the training. The small weights are frozen between the first and second phase with a threshold of $\alpha=0.1$. The second phase uses a base learning rate of $10^{-3}$ and no regularization for 10000 iterations.

\subsection{SHO}

To generate the SHO dataset, we sample 1000 datapoints values for $\omega^2$ and generate time series time series $\{\mathbf{x}_t\}_{t=0}^{T_x-1}$ and $\{\mathbf{y}_t\}_{t=0}^{T_y}$ for each $\omega^2$. The input series is propagated for $T_x=500$ time steps with a time step of $\Delta t=0.1$. The output series is propagated for $T_y=25$ time steps with the same time step.

The dynamics encoder is the same architecture as used in the kinematics experiment. Due to the greater number of time steps that the system needs to propagate, the EQL network does not duplicate the activation functions for all functions. The functions in each hidden layer consist of:
\[ [1 (\times 2), g (\times 2), g^2 , \sin (2\pi g) , e^g , 10g_1*g_2 (\times 2)] \]
The ReLU network that is trained in place of the EQL network for comparison consists of four hidden layers with 50 units each and ReLU activation functions.

We use three phases of training, where the first phase uses a learning rate of $10^{-2}$ and a regularization weight of $\lambda=4\times10^{-5}$ for a total of 2000 iterations. The system starts training on $T_y=1$ time steps for the first 500 time steps and then add 2 more time steps every 500 iterations for a total of $T_y=7$ time steps. In the second phase of training, we increase the number of time steps to $T_y=25$, decrease the base learning rate to $2\times10^{-3}$, and increase the regularization weight to $\lambda=2\times10^{-4}$. The small weights are frozen between the second and third phase with a threshold of $\alpha=0.01$. The third and final phase of training uses a base learning rate of $10^{-3}$ and no regularization.

\section{Additional MNIST Arithmetic Data}

The results presented in Figure \ref{fig:mnist_result} and Table \ref{table:mnist} are drawn from one of several trials, where each in each trial the network is trained from a different random initialization of the network weights. Due to the random initialization, the EQL does not reach the same equation every time. Here we present results from additional trials to demonstrate the variability in the system's behavior as well as the system's robustness to the random initializations.

The experimental details are described in Section \ref{sec:mnist} where digits $\chi_{1,2}$ are drawn from the entire MNIST training dataset. We refer to the results shown in Figure \ref{fig:mnist_result} and Table \ref{table:mnist} as Trial 1.

\begin{figure}[tbp]
\centering
    \includegraphics[width=0.8\columnwidth,trim=3 0 0 2,clip]{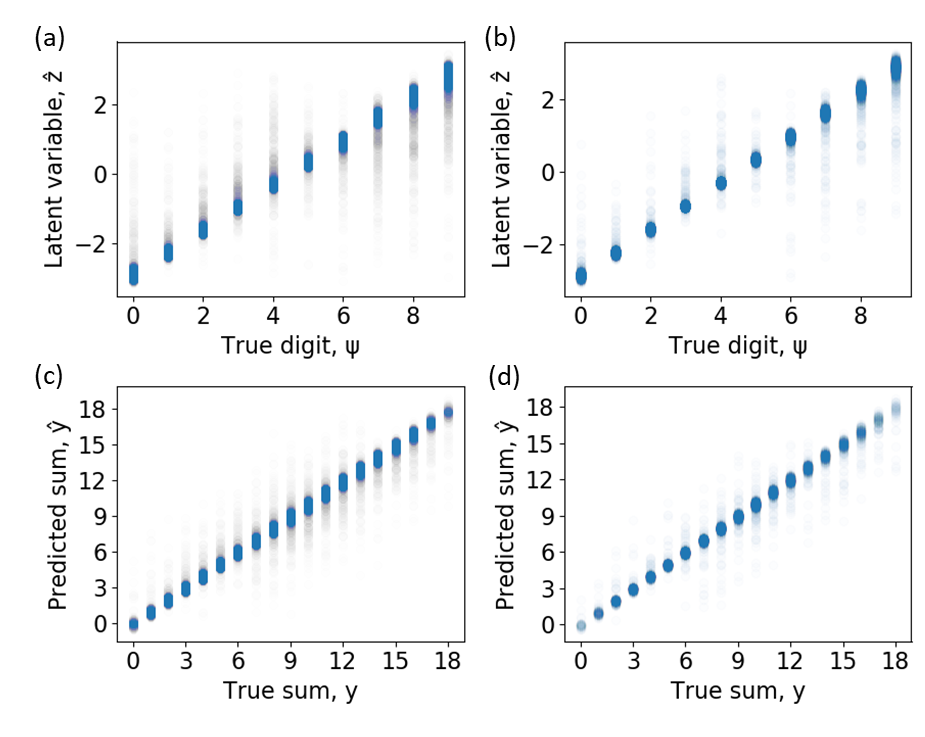}
	\caption{The ability of the encoder to differentiate between digits as measured by the latent variable $z$ versus the true digit $\psi$ for digits $\chi$ drawn from the MNIST (a) training dataset and (b) test dataset. The ability of the entire architecture to fit the label $y$ as measured by the predicted sum $\hat{y}$ versus the true sum $y$ for digits $\chi$ drawn from the MNIST (c) training dataset and (d) MNIST test dataset.}
	\label{fig:mnist_trial1}
\end{figure}

The results for Trial 2 are shown in Figure \ref{fig:mnist_trial1}. Similar to Trial 1, Trial 2 produces a linear relationship between the true digit $\phi$ and the latent variable $z$, although there is a positive instead of negative correlation. As previously mentioned, there is no bias placed on the latent variable $z$ so whether there is a positive or negative correlation is arbitrary and depends on the random initialization of the weights. The trained architecture produced the following expression from the EQL network:

\begin{equation}
\hat{y} = 1.565 z_1 + 1.558 z_2 + 9
\label{eq:mnist2}
\end{equation}

\noindent Note the positive coefficients in (\ref{eq:mnist2}) which reflects the positive correlation shown in Figure \ref{fig:mnist_trial1}(a-b). As shown in Figure \ref{fig:mnist_trial1}(c-d), the network is still able to accurately predict the sum $y$.

\begin{figure}[tbp]
\centering
    \includegraphics[width=0.8\columnwidth,trim=3 0 0 2,clip]{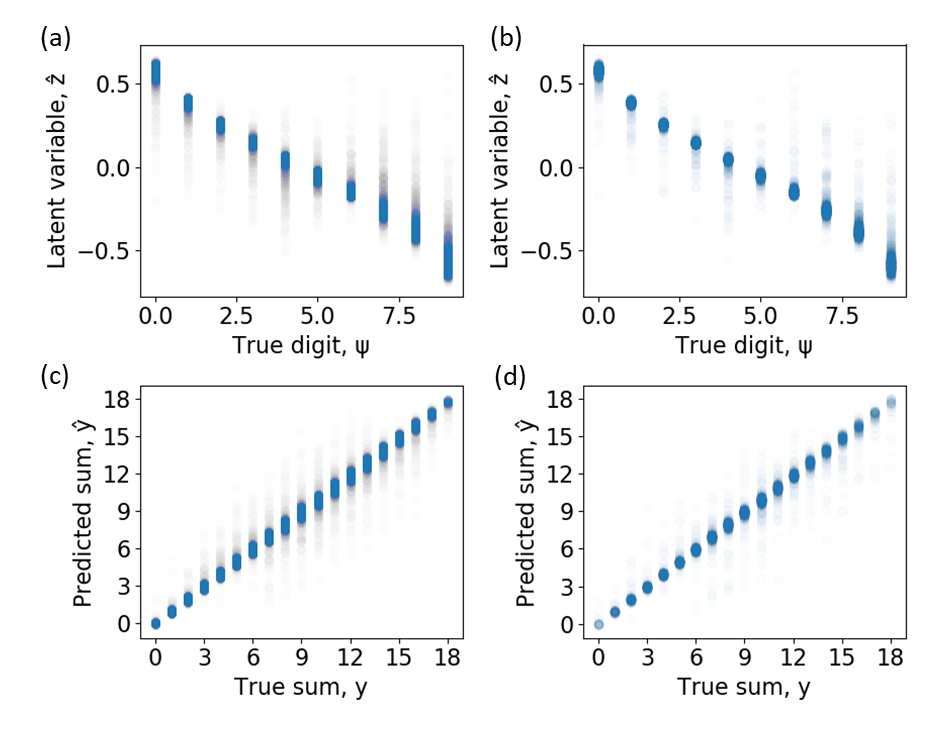}
	\caption{The ability of the encoder to differentiate between digits as measured by the latent variable $z$ versus the true digit $\psi$ for digits $\chi$ drawn from the MNIST (a) training dataset and (b) test dataset. The ability of the entire architecture to fit the label $y$ as measured by the predicted sum $\hat{y}$ versus the true sum $y$ for digits $\chi$ drawn from the MNIST (c) training dataset and (d) MNIST test dataset.}
	\label{fig:mnist_trial2}
\end{figure}

The results for Trial 3 are shown in Figure \ref{fig:mnist_trial2}. Note that in this case, the relationship between $\phi$ and $z$ is no longer linear. However, the encoder still finds a one-to-one mapping between $\phi$ and $z$, and the EQL network is still able to extract the information from $z$ such that it can predict the correct sum as shown in Figure \ref{fig:mnist_trial2}(c-d).

The equation found by the EQL network is:

\begin{equation}
\hat{y} = -4.64\sin(2.22 z_1) - 4.63 \sin(2.21 z_2) + 9
\label{eq:mnist3}
\end{equation}

\noindent This is consistent with the insight that the curve in Figure \ref{fig:mnist_trial2}(a-b) represents an inverse sine function. Thus, (\ref{eq:mnist3}) is first inverting the transformation from $\phi$ to $z$ to produce a linear mapping and then adding the two digits together. So while the EQL network does not always give the exact equation we expect, we can still gain insight into the system from analyzing the latent variable and the resulting equation.

\section*{Acknowledgment}

We would like to acknowledge Joshua Tenenbaum, Max Tegmark, Jason Fleischer, Alexander Alemi, Jasper Snoek, Stjepan Picek, Rumen Dangovski, and Ilan Mitnikov for fruitful conversations. This research is sponsored in part by the Army Research Office and under Cooperative Agreement Number W911NF-18-2-0048, by the Department of Defense through the National Defense Science \& Engineering Graduate Fellowship (NDSEG) Program, by the MIT–SenseTime Alliance on Artificial Intelligence, by the Defense Advanced Research Projects Agency (DARPA) under Agreement No. HR00111890042.
Research was also sponsored in part by the United States Air Force Research Laboratory and was accomplished under Cooperative Agreement Number FA8750-19-2-1000. The views and conclusions contained in this document are those of the authors and should not be interpreted as representing the official policies, either expressed or implied, of the United States Air Force or the U.S. Government. The U.S. Government is authorized to reproduce and distribute reprints for Government purposes notwithstanding any copyright notation herein.

\ifCLASSOPTIONcaptionsoff
  \newpage
\fi
%

\bibliographystyle{ieeetran}
\bibliography{sym}


\end{document}